\documentclass{article}
\usepackage{spconf,amsmath,amssymb,graphicx}
\usepackage{mathtools}
\usepackage{subfig}
\usepackage{siunitx}
\usepackage{bmpsize}
\usepackage[abs]{overpic}
\usepackage{adjustbox}
\usepackage{booktabs,array}
\usepackage{multirow}
\usepackage{balance}
\renewcommand{\thetable}{\Roman{table}}

\usepackage[usenames,dvipsnames]{color}
\usepackage[colorlinks=true,linkcolor=NavyBlue, citecolor=NavyBlue, urlcolor=NavyBlue]{hyperref}

\makeatletter
\renewcommand\footnotesize{%
	\@setfontsize\footnotesize\@ixpt{9.7}%
	\abovedisplayskip 8\p@ \@plus2\p@ \@minus4\p@
	\abovedisplayshortskip \z@ \@plus\p@
	\belowdisplayshortskip 4\p@ \@plus2\p@ \@minus2\p@
	\def\@listi{\leftmargin\leftmargini
		\topsep 4\p@ \@plus2\p@ \@minus2\p@
		\parsep 2\p@ \@plus\p@ \@minus\p@
		\itemsep \parsep}%
	\belowdisplayskip \abovedisplayskip
}
\makeatother

\title{Retrospective correction of Rigid and Non-Rigid MR motion artifacts using GANs}
\name{Karim Armanious\textsuperscript{1,2}, Sergios Gatidis\textsuperscript{2}, Konstantin Nikolaou\textsuperscript{2}, Bin Yang\textsuperscript{1}, Thomas K\"ustner\textsuperscript{1,2,3}}
\address{\textsuperscript{1}University~of~Stuttgart~,~Institute~of~Signal~Processing~and~System~Theory,~Stuttgart,~Germany\\
	\textsuperscript{2}University~of~T\"ubingen,~Department~of~Radiology,~T\"ubingen,~Germany\\
	\textsuperscript{3}King's~College~London,~Biomedical~Engineering~Department,~London,~England
}
\begin{document}
%
\maketitle
\begin{abstract}
Motion artifacts are a primary source of magnetic resonance (MR) image quality deterioration with strong
repercussions on diagnostic performance. Currently, MR motion correction is 
carried out either prospectively, with the help of motion tracking systems, or 
retrospectively by mainly utilizing computationally expensive iterative algorithms. In this paper, 
we utilize a new adversarial framework, titled MedGAN, for the joint retrospective
correction of rigid and non-rigid motion artifacts in different body regions and without the need for a reference image. MedGAN utilizes a unique combination of non-adversarial losses and a new generator architecture to capture
the textures and fine-detailed structures of the desired artifact-free MR images.
Quantitative and qualitative comparisons with other adversarial techniques
have illustrated the proposed model performance.
\end{abstract}
\section{Introduction}
\label{sec:intro}

Magnetic Resonance Imaging (MRI) is a cornerstone of modern medical diagnostic techniques. It is used to acquire detailed anatomical and physiological information of different organs and processes. This has strong relevance in the field of oncology, specifically the diagnosis, staging and follow-up of tumours. Nevertheless, MRI often suffers from motion-related artifacts due to the physiological motion (breathing, cardiac motion, peristaltic) during longer examinations. These motion artifacts manifest in the images as blurring, aliasing and stretching of the underlying anatomical structures which degrade the information content of the acquired MR images even hindering diagnostic evaluation. 

Motion can be categorized into two main categories, rigid and non-rigid motion. Rigid artifacts are caused by global deformations caused by the bulk motion of a whole body part, e.g. movements of the arms or head. Non-rigid artifacts  are more subtle. They are local deformations arising from involuntary patient motion, e.g. respiratory or cardiac motion.  

The simplest and most effective strategy of mitigating motion-artifacts is intercepting patient motion in the first place, e.g. by scanning under breath hold or if necessary under sedation. However, this approach has obvious limitations regarding patient safety and comfort \cite{1}. The majority of available approaches for motion correction are concerned with real-time prospective correction during the MR acquisition process. This relies on internal or external tracking systems which are used to track the patient's motion and thus trigger or guide the acquisition procedure. Examples of utilized tracking systems include cameras \cite{2}, active markers \cite{3}, MR navigators \cite{4}, respiratory belts \cite{5} and ultrasound signals \cite{6}. However, these approaches may result in prolonged or unpredictably scan time (in case of triggering). Moreover, motion artifacts can still occur despite these techniques, especially when motion is non-rigid.

Retrospective correction of motion-artifacts takes place after the MR acquisition procedure. This can be carried out by utilizing tracking devices analogous to prospective methods \cite{7}. Additionally, iterative methods such as optimizing the entropy of spatial gradients \cite{8} or MR autofocusing techniques \cite{9,10} can also be utilized. However, they suffer from being computationally intensive as well as difficulty in coping with complex non-rigid motion artifacts \cite{11}. 

Recent advances in deep learning algorithms, more specifically convolutional neural networks (CNNs), have led to breakthrough results in several medical applications such as semantic segmentation \cite{12,13}, lesion classification \cite{14}, anomaly detection \cite{15} and data augmentation \cite{16}. These advances were extended to the correction of MR motion artifacts by means of variational autoencoders for the correction of very mild rigid motion artifacts in the head region \cite{17}. In our previous work, a novel framework based on generative adversarial networks (GANs), named MedGAN, was proposed for the correction of severe rigid motion artifacts in the head region achieving state-of-the-art results \cite{18}.

In this work, we extend our previous MedGAN framework for the simultaneous correction of severe rigid and non-rigid motion artifacts from several body regions. This is achieved will being independent from any required surrogate motion model or reference image. Moreover, we compare the results quantitatively and qualitatively against several state-of-the-art GAN based methods. Additionally, we illustrate the advantage of the joint correction of rigid and non-rigid motion artifacts by comparing against an identical model trained solely on a single type of motion artifacts.

%
%
%

\section{Methods}
\label{sec:format}

\begin{figure}[!t]
	\centering
	
	\includegraphics[width=0.5\textwidth]{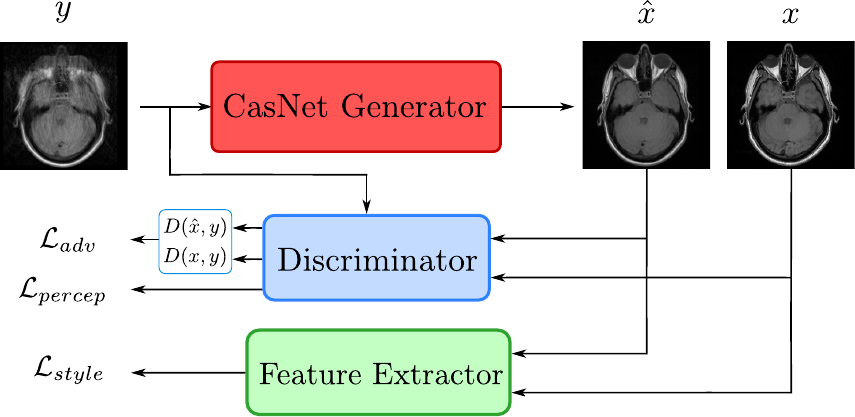}
	
	\caption{An overview of the MedGAN framework utilized for the retrospective correction of MR motion artifacts.}
	\label{2}
\end{figure}

MedGAN is based on conditional generative adversarial networks (cGANs) with the inclusion of additional non-adversarial losses and a novel generator architecture \cite{18}. An overview of the proposed framework is presented in Fig.~\ref*{2}.

\subsection{Conditional Generative Adversarial Networks}

A cGAN consists of two main components, a conditional generator network $G$ and a discriminator network $D$ \cite{22}. For MR motion correction, the generator receives as input a motion corrupted MR image $y$ and maps it into a corresponding denoised MR image, $\hat{x} = G(y)$. The discriminator, however, receives both the generator output $\hat{x}$ as well as the ground-truth target domain image $x$ as inputs. It then acts as a binary classifier attempting to distinguish which of the inputs belong to the real target distribution, $D(x,y) = 1$, and which is the fake synthetic output from the generator, $D(\hat{x},y) = 0$. Both networks are pitted against each other in competition in which the generator attempts to produce realistic corrected MR images, thus fooling the discriminator network, while the discriminator seek to improve its classification performance and thus avoid to be fooled. This can be represented by the adversarial loss function $\mathcal{L}_{\small\textrm{adv}}$ which is formulated as the following min-max optimization task:

\begin{equation}
\min_{G} \max_{D} \mathcal{L}_{\small\textrm{adv}} = \mathbb{E}_{x,y} \left[\textrm{log} D(x,y) \right] + \mathbb{E}_{\hat{x},y} \left[\textrm{log} \left( 1 - D\left(\hat{x},y\right) \right) \right]
\end{equation}

\subsection{Non-Adversarial Losses}

Results from cGAN frameworks often exhibit blurriness and loss of fine-detailed structures which is especially critical in the context of medical images. Thus, MedGAN incorporates additional non-adversarial loss components to counteract this phenomenon. The first loss component is the perceptual loss which is constructed to ensure global consistency while capturing the perceptual quality of human judgement. This is achieved by utilizing the discriminator as a trainable feature extractor. As such, it is used to calculate the mean absolute error (MAE) between the intermediate features of the synthetic output $\hat{x}$ versus those of the desired target image $x$:
 
 \begin{equation}
 P_i \left(\hat{x},x\right) = \lVert{D_i\left(\hat{x},y\right) - D_i\left(x,y\right)}\rVert_1
 \end{equation}
 where $D_i$ represents the extracted feature representation of the $i^{\textrm{th}}$ layer of the discriminator. The generator then minimizes the total perceptual loss which is calculated as a weighted sum of the MAE for each layer of the discriminator:
 
 \begin{equation}
 \mathcal{L}_{\small\textrm{percep}} = \sum_{i = 0}^{L} \lambda_{pi} P_i \left(\hat{x},x\right)
 \label{7}
 \end{equation}
 where $\lambda_{pi} > 0 $ is the weight for the $i^{\textrm{th}}$ layer and $L$ is the total number of the discriminator hidden layers. Additionally, the perceptual loss incorporates internally an additional pixel reconstruction component by computing the MAE of the raw image inputs ($i=0$).
 
 Inspired by recent advances in the field of neural style transfer \cite{1920}, an additional style reconstruction loss is utilized to enhance the textures and fine details of the resultant denoised MR images. This loss is achieved by utilizing a pre-trained feature extractor network, e.g. a VGG-19 network pre-trained on ImageNet classification task \cite{21}, for the calculation of the feature correlations over the spatial extent represented by the Gram matrix $G_i(x)$. The elements of this matrix are:
 
 \begin{equation}
 G_i(x)_{m,n} = \frac{1}{h_i w_i d_i} \sum_{h = 1}^{h_i} \sum_{w = 1}^{w_i} V_{i}(x)_{h,w,m} V_{i}(x)_{h,w,n}
 \end{equation}
 where $V_i(x)$ represents the extracted feature representation in the $i^{\textrm{th}}$ layer of the pre-trained network with $h_i$, $w_i$, $d_i$ the representation height, width and spatial depth respectively. Finally, the style reconstruction loss to be minimized is the weighted average of the squared Frobenius norm of the deviation between the Gram matrix of the denoised output $\hat{x}$ and that of the ground truth target images $x$:
 
 \begin{equation}
 \mathcal{L}_{\small\textrm{style}} = \sum_{i = 1}^{B} \lambda_{si} \frac{1}{4 d_i^2} \lVert{G_i\left(\hat{x}\right) - G_i\left(x\right)}\rVert_F^{2}
 \end{equation}
 where $\lambda_{si} > 0 $ is the weight of the $i^{\textrm{th}}$ of the pre-trained network and $B$ is the total number of layers.
 
 \subsection{Network Architecture}
 
 For the discriminator network, we utilize a patch discriminator architecture, identical to that introduced in \cite{22}. It divides the input images into smaller patches, classifies each patch as either real or fake and finally it averages out the score of all image patches. 
 
 For the generator architecture, we similarly utilize the U-net architecture presented in \cite{22} as our basic building block. It is a fully convolutional encoder-decoder architecture which incorporates batch normalization and skip-connections to maintain low-level spatial information through the bottleneck layer. Additionally, to increase the generative capacity of our model and to ensure a crisp and high-resolution output, we concatenate three U-nets in an end-to-end manner into the CasNet architecture, see \cite{18,29} for more details. As a result, the motion correction task is distributed over the more extensive network, and the synthetically corrected MR images are progressively enhanced as they go through the different encoder-decoder pairs compared to using the conventional U-net architecture.

%

\section{Experimental Evaluations}
\label{sec:pagestyle}


We evaluated our proposed MedGAN model for the retrospective correction of MR motion artifacts on a dataset from 17 anonymized volunteers in the head, pelvis and abdomen regions. The datasets were acquired using a 2D multi-slice T1-weighted spin echo (SE) sequence on a clinical 3 Tesla MR scanner \cite{23}. Each volunteer was scanned twice, once under resting (head, pelvis) and breath-hold conditions (abdomen), and another while inducing rigid motion artifacts through the movements of the head and hips and non-rigid artifacts through respiratory motion. 2-dimensional slices were extracted and scaled to a matrix size of $256 \times 256$ pixels. 

Motion-free MR scans were aligned and paired with their corresponding motion-corrupted scans from the same volunteer. The training dataset for the rigid motion artifacts, from the head and pelvis regions, consists of 980 paired images from 14 patients and 105 images from 3 patients for validation, each containing a motion-free and the corresponding motion-corrupted image. For the non-rigid respiratory artifacts, in the abdomen region, 420 paired trained images were used from 14 patients and 90 paired images for validation from 3 patients. 


To showcase the capabilities of our proposed model, we compare the results qualitatively and quantitatively against those achieved by other state-of-the-art GAN techniques. Specifically, we implement and train the pix2pix framework \cite{22}, which combines a cGAN with a pixel reconstruction loss, and the ID-cGAN framework \cite{24}, a state-of-the-art denoising method. The frameworks mentioned above and our proposed MedGAN model were trained for correction of motion artifacts from a single body region. Additionally, another instance of MedGAN, with this mentioned as MedGAN-joint, was trained for the joint motion correction from all available body regions. All models utilized the same hyperparameters, network architectures (except for CasNet generator architecture) and were trained for 100 epochs using a single NVIDIA Titan X GPU.

For the quantitative comparisons, we utilize the Structural Similarity Index (SSIM)
\cite{25}, Visual Information Fidelity (VIF) \cite{26}, Universal Quality Index (UQI) \cite{27} and Learned Perceptual Image Patch Similarity (LPIPS) \cite{28} as evaluation metrics.

\section{Results and Discussion}
\label{sec:typestyle}

\begin{figure*}[!t]
	\begin{minipage}[t]{1.0\linewidth}
		\centering
		\vspace{7mm}
		\begin{minipage}[t]{0.175\linewidth}
			\centering
			\begin{overpic}[width=0.837\textwidth]%
				{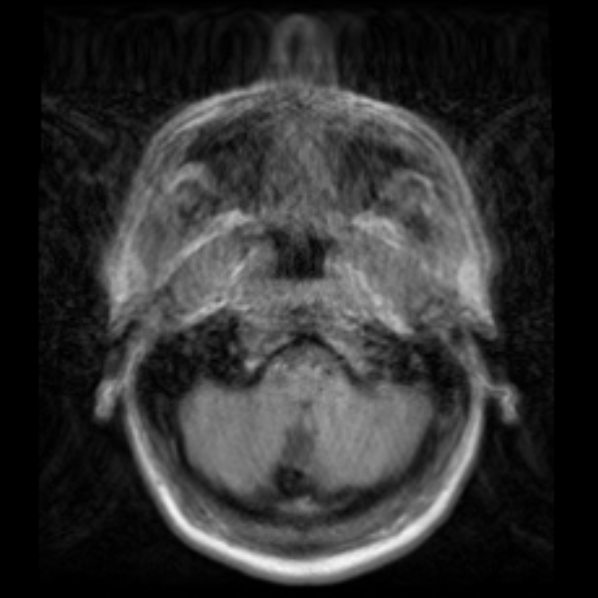}
				\centering
				\put(26,83){Input}
			\end{overpic}	
		\end{minipage}%
		\begin{minipage}[t]{0.645\linewidth}
			\centering
			\begin{overpic}[width=0.227\textwidth]%
				{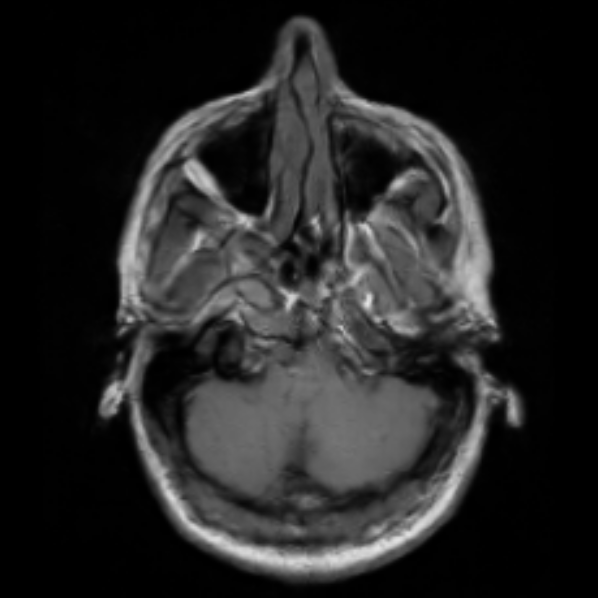}
				\centering
				\put(22,83){pix2pix}
			\end{overpic}
			\begin{overpic}[width=0.227\textwidth]%
				{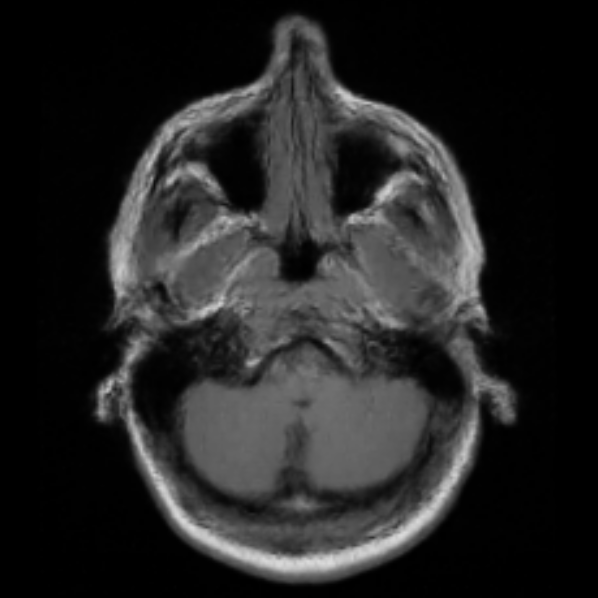}
				\centering
				\put(16,83){ID-cGAN}
			\end{overpic}
			\begin{overpic}[width=0.227\textwidth]%
				{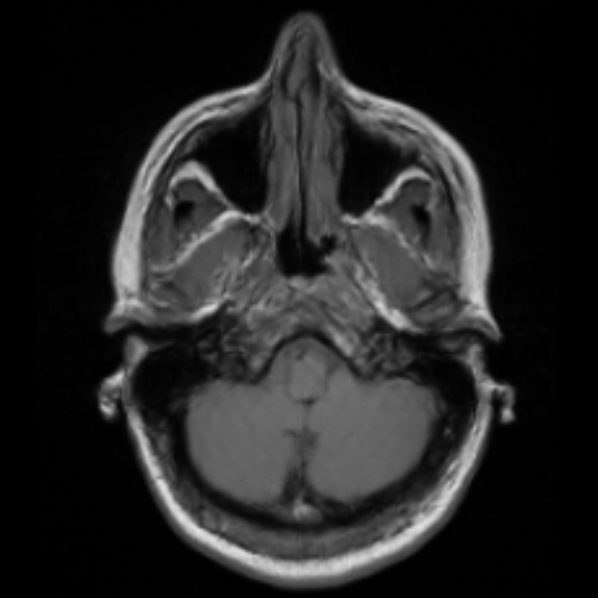}
				\centering
				\put(16,83){MedGAN}
			\end{overpic}
			\begin{overpic}[width=0.227\textwidth]%
				{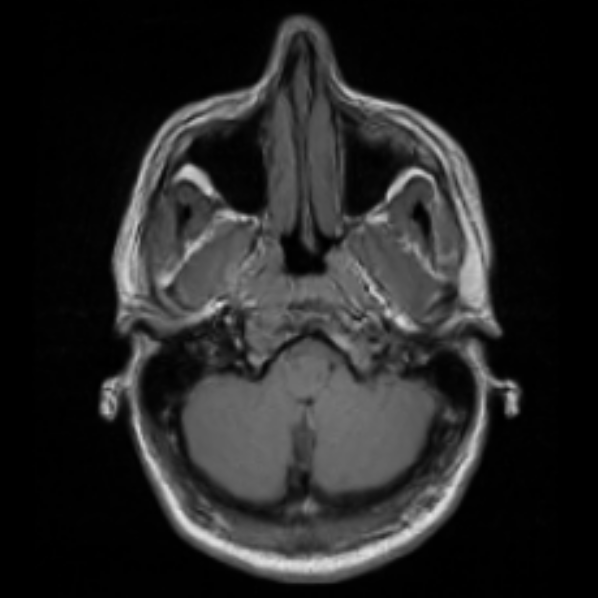}
				\centering
				\put(6,83){MedGAN-joint}
			\end{overpic}
		\end{minipage}
		\begin{minipage}[t]{0.160\linewidth}
			\centering
			\begin{overpic}[width=0.916\textwidth]%
				{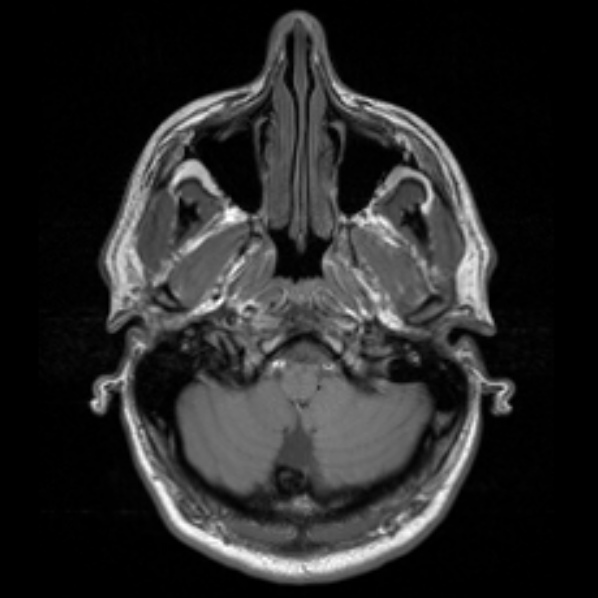}
				\centering
				\put(22,83){Target}
			\end{overpic}		
		\end{minipage}\\
	
		(a) Head motion correction (rigid)
	\end{minipage}\\
	
	\vspace{9mm}

	\begin{minipage}[t]{1.0\linewidth}
		\centering
		\begin{minipage}[t]{0.175\linewidth}
			\centering
			\begin{overpic}[width=0.837\textwidth]%
				{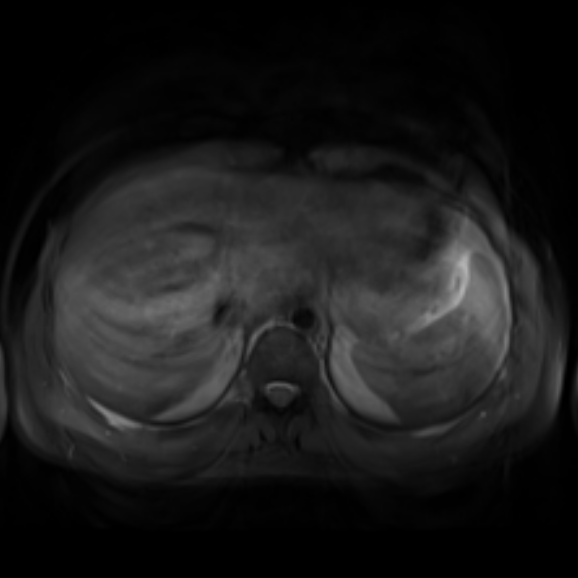}
				\centering
				\put(26,83){Input}
			\end{overpic}	
		\end{minipage}%
		\begin{minipage}[t]{0.645\linewidth}
			\centering
			\begin{overpic}[width=0.227\textwidth]%
				{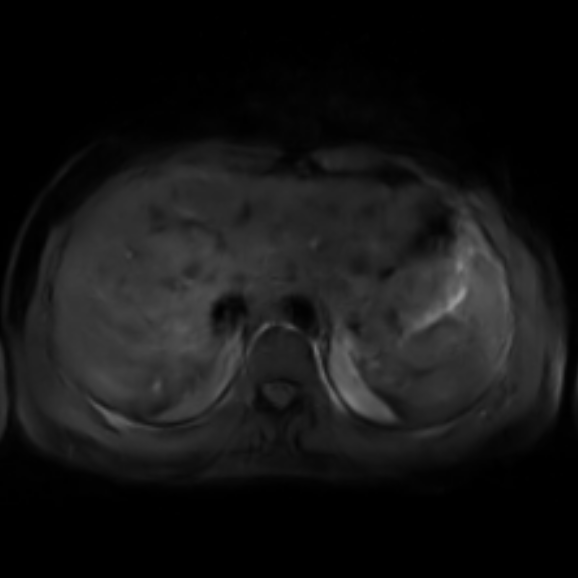}
				\centering
				\put(22,83){pix2pix}
			\end{overpic}
			\begin{overpic}[width=0.227\textwidth]%
				{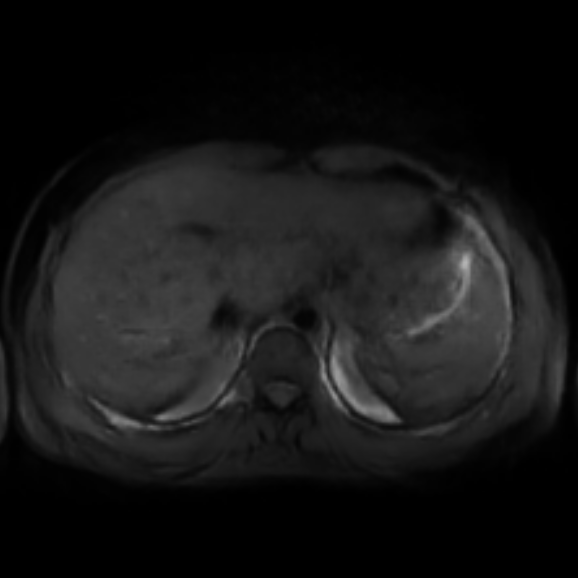}
				\centering
				\put(16,83){ID-cGAN}
			\end{overpic}
			\begin{overpic}[width=0.227\textwidth]%
				{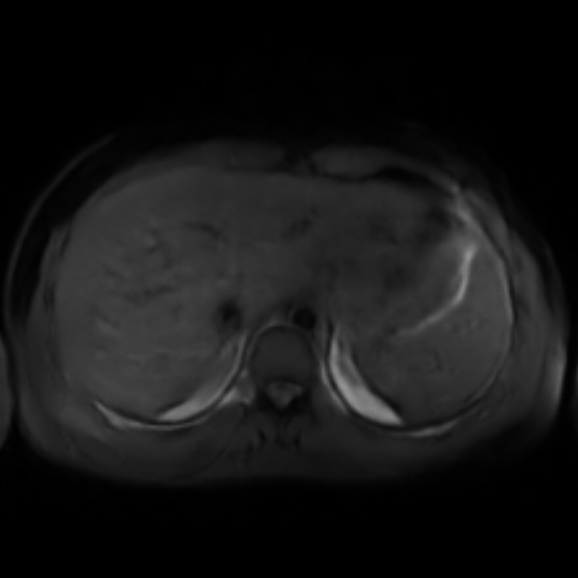}
				\centering
				\put(16,83){MedGAN}
			\end{overpic}
			\begin{overpic}[width=0.227\textwidth]%
				{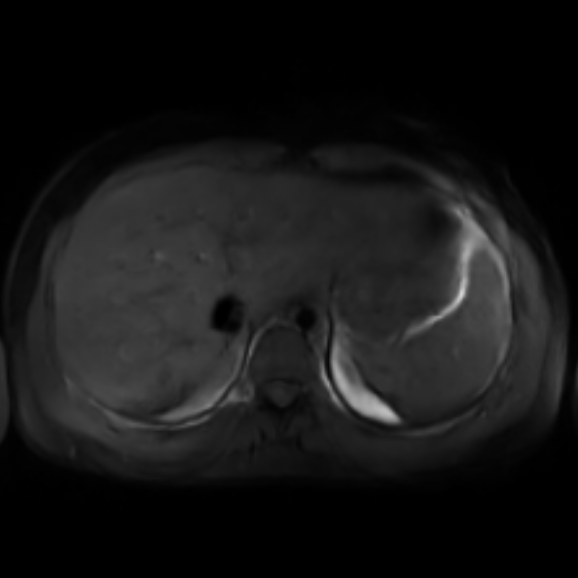}
				\centering
				\put(6,83){MedGAN-joint}
			\end{overpic}
		\end{minipage}
		\begin{minipage}[t]{0.160\linewidth}
			\centering
			\begin{overpic}[width=0.916\textwidth]%
				{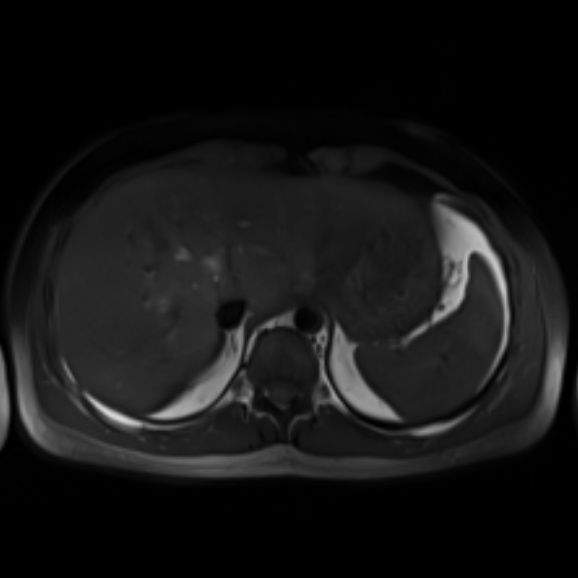}
				\centering
				\put(22,83){Target}
			\end{overpic}		
		\end{minipage}\\

		(b) Abdomen motion correction (non-rigid)
	\end{minipage}
	
	\vspace{9mm}
	
	\begin{minipage}[t]{1.0\linewidth}
		\centering
		\begin{minipage}[t]{0.175\linewidth}
			\centering
			\begin{overpic}[width=0.837\textwidth]%
				{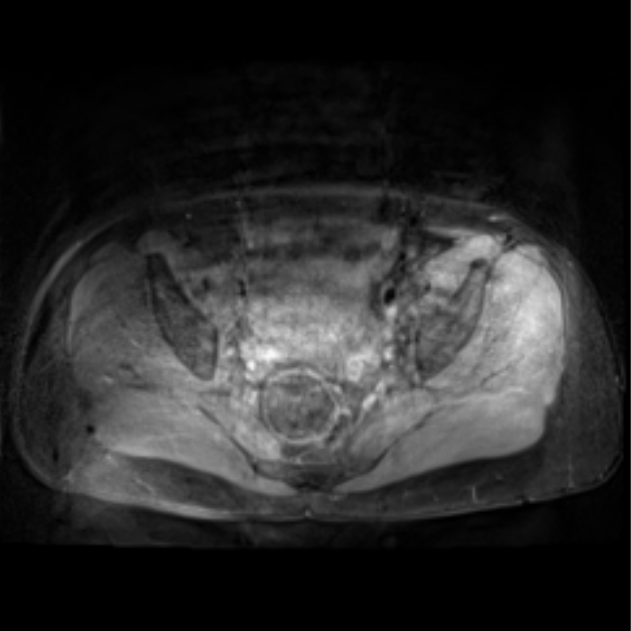}
				\centering
				\put(26,83){Input}
			\end{overpic}	
		\end{minipage}%
		\begin{minipage}[t]{0.645\linewidth}
			\centering
			\begin{overpic}[width=0.227\textwidth]%
				{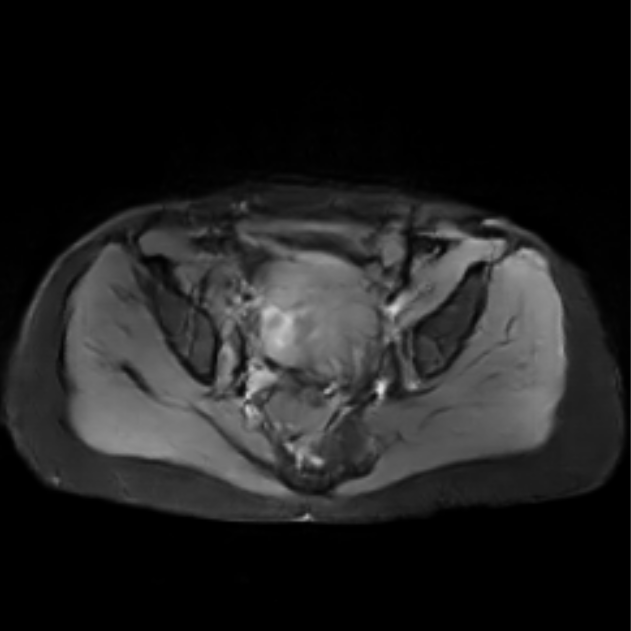}
				\centering
				\put(22,83){pix2pix}
			\end{overpic}
			\begin{overpic}[width=0.227\textwidth]%
				{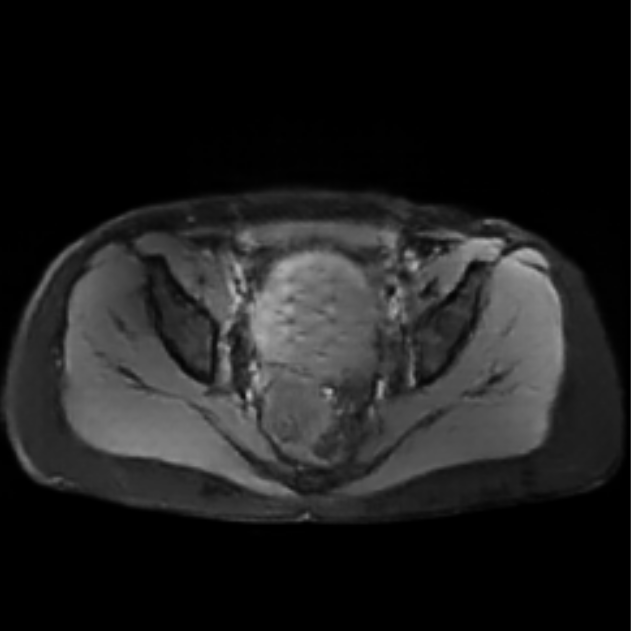}
				\centering
				\put(16,83){ID-cGAN}
			\end{overpic}
			\begin{overpic}[width=0.227\textwidth]%
				{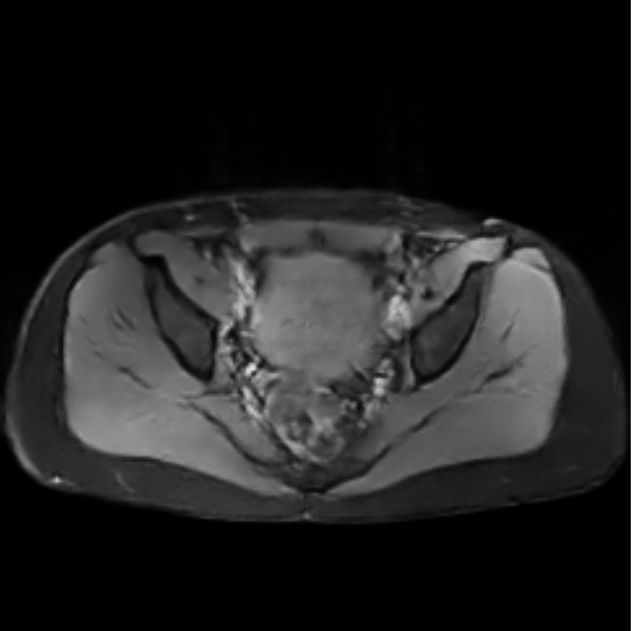}
				\centering
				\put(16,83){MedGAN}
			\end{overpic}
			\begin{overpic}[width=0.227\textwidth]%
				{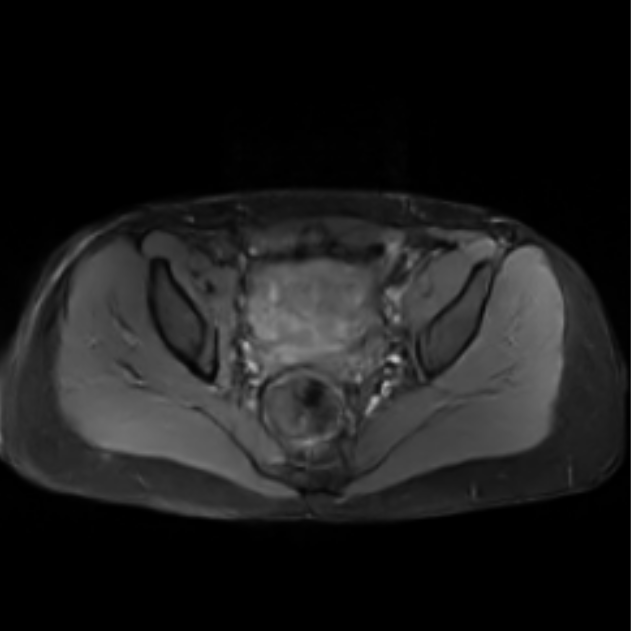}
				\centering
				\put(6,83){MedGAN-joint}
			\end{overpic}
		\end{minipage}
		\begin{minipage}[t]{0.160\linewidth}
			\centering
			\begin{overpic}[width=0.916\textwidth]%
				{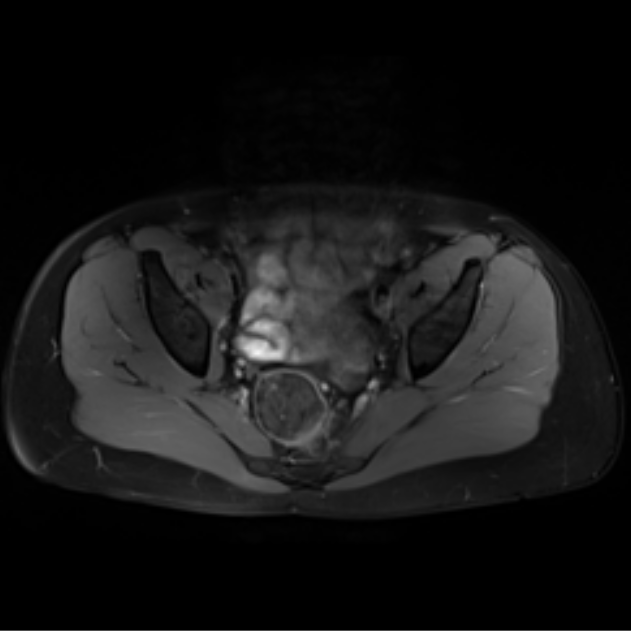}
				\centering
				\put(22,83){Target}
			\end{overpic}		
		\end{minipage}\\

		(c) Pelvis motion correction (rigid)
	\end{minipage}
	\caption{Comparison of the results of MedGAN and different state-of-the-art techniques for the correction of MR motion artifacts from different body regions.}
	\label{5}
\end{figure*}

\begin{table*}[h!]
	\caption{Quantitative comparison with state-of-the-art frameworks for different body regions}
	\centering
	\label{t2}
	\scriptsize
	\bgroup
	\def\arraystretch{1.7}
	\begin{adjustbox}{width=1.0\textwidth}
		\begin{tabular}{r|cccc|cccc|cccc}
			\noalign{\smallskip} \hline \hline \noalign{\smallskip}
			\multirow{2}{*}{Method} & \multicolumn{4}{c}{(a) Head motion correction} & %
			\multicolumn{4}{c}{(b) Abdomen motion correction} & \multicolumn{4}{c}{(c) Pelvis motion correction}\\
			& SSIM & VIF & UQI & LPIPS & SSIM & VIF & UQI & LPIPS & SSIM & VIF & UQI & LPIPS\\
			\hline
			pix2pix & 0.8238 & 0.3465 & 0.5464 & 0.2728 & 0.8160 & 0.3366 & 
			0.6815 & 0.3064 & 0.6690 & 0.1843 & 0.6540 & 0.5683 \\
			
			ID-cGAN & 0.8329 & 0.3630 & 0.5795 & 0.2633 & 0.8307 & 0.3718 
			& 0.6955 & 0.2787 & 0.6770 & 0.1892 & 0.6339 & 0.5586\\
			
			MedGAN & \textbf{0.8369} & 0.3664 & 0.5821 & \textbf{0.2202} & 0.8321 & \textbf{0.3779}
			& \textbf{0.7523} & 0.2508 & 0.6763 & 0.1895 & 0.6901 & 0.5414\\
			
			
			MedGAN-joint & 0.8314 & \textbf{0.3744} & \textbf{0.6571} & 0.2291 & \textbf{0.8335} & 0.3734 
			& 0.7516 & \textbf{0.2475} & \textbf{0.6854} & \textbf{0.1951} & \textbf{0.6921} & \textbf{0.5357}\\
			
			\noalign{\smallskip} \hline \noalign{\smallskip}
		\end{tabular}
	\end{adjustbox}
	\egroup
\end{table*}

The results for the qualitative and quantitative comparison between MedGAN and other state-of-the-art adversarial techniques are presented in Fig.~\ref*{5} and Table~I, respectively. The pix2pix framework resulted in the worst quantitative performance across the utilized evaluation metrics. This is also reflected in the qualitative comparison for the head, abdomen and pelvis regions. Although it succeeded in providing sharp denoised images with no blurriness, the resultant MR images
by pix2pix lacked homogeneity and global consistency with unrealistic biological structures. ID-cGAN outperformed pix2pix quantitatively by producing more globally structured images. Nevertheless, from a qualitative perspective, ID-cGAN resulted in blurred results with a lack of sharpness and fine details. 

The proposed MedGAN model, trained for the motion correction of specific body regions, surpassed the previously mentioned techniques from both the qualitative and quantitative aspects. This is highlighted by the visual fidelity and structural consistency presented in Fig.~\ref*{5}. However, the performance of MedGAN can be further enhanced by training jointly on all available body regions containing both rigid and non-rigid motion artifacts. As a result of the learned correlation between different body regions, MedGAN-joint results in sharper motion corrected images with a higher level of details resembling that of the target motion-free images.

To summarize, in this work we present MedGAN as a method for the retrospective correction of rigid and non-rigid MR motion artifacts. The above results highlight the capabilities of MedGAN for producing near-realistic MR images from severely deteriorated scans. However, MedGAN is not without limitations. During the motion correction procedure, relevant diagnostic information can be lost and needs further detailed investigation. At this stage, we do not aim to utilize MedGAN for diagnosis but rather for technical post-processing tasks which require globally consistent image properties. Reaching diagnostic quality will be explored in future studies. Examples of such tasks include using already acquired but corrupted MR scans for segmentation, organ volume estimation, attenuation correction for PET/MR scans and automatic detection of different anatomic regions.

\section{Conclusion}
\label{sec:ref}

MedGAN, a novel adversarial framework, is an effective solution for the retrospective joint correction of both rigid and non-rigid motion artifacts without the necessity of motion tracking devices. MedGAN combines an adversarial framework with a combination of non-adversarial losses and a novel generator architecture to produce near-realistic results. Quantitative and qualitative comparisons with different adversarial techniques for translation and denoising showcased MedGAN's superior performance in the task of motion correction.

In the future, we plan to investigate the performance of MedGAN on complex-valued data by including phase information in addition to the currently utilised magnitude information on a larger training dataset. Moreover, we plan to extend our model to 3-dimensional space with further subjective evaluations of performance by experienced radiologists. 


\newpage
\bibliographystyle{IEEEbib}
\balance
{\footnotesize
	\bibliography{refs2}}

\end{document}